\documentclass[conference]{IEEEtran}
\IEEEoverridecommandlockouts
\usepackage{cite}
\usepackage{comment}
\usepackage{amsmath,amssymb,amsfonts}
\usepackage{algorithmic}
\usepackage{graphicx}
\usepackage{textcomp}
\usepackage[table, svgnames, dvipsnames]{xcolor}
\usepackage{dirtree}
\usepackage{enumitem}
\usepackage{booktabs}
\usepackage{soul}
\usepackage{listings}
\usepackage{pifont}
\usepackage{multirow}
\usepackage{hyperref}
\usepackage{color, colortbl}	
\definecolor{LightCyan}{rgb}{0.88,1,1}

\usepackage[a4paper, total={184mm,239mm}]{geometry}
\def\BibTeX{{\rm B\kern-.05em{\sc i\kern-.025em b}\kern-.08em
    T\kern-.1667em\lower.7ex\hbox{E}\kern-.125emX}}
\usepackage{cleveref}
\definecolor{gray}{rgb}{0.4,0.4,0.4}
\definecolor{purple}{rgb}{0.128,0,0.128}
\definecolor{darkblue}{rgb}{0.0,0.0,0.6}
\definecolor{cyan}{rgb}{0.0,0.6,0.6}

\lstdefinelanguage{XML}
{
  morestring=[b]",
  morestring=[s]{>}{<},
  morecomment=[s]{<?}{?>},
  stringstyle=\color{darkblue},
  identifierstyle=\color{darkblue},
  keywordstyle=\color{cyan},
  morekeywords={fmiModelDescription,Int32,causality,name}%
}

\newcommand{\tlm}{SystemC TLM}

\renewcommand{\baselinestretch}{0.975}
\usepackage{tikz}
\usepackage{textcomp}
\usepackage[doipre={DOI:~}]{uri}
\usepackage{lipsum}
\newcommand\copyrighttext{%
  \footnotesize \textcopyright 2025 IEEE. Personal use of this material is permitted.  Permission from IEEE must be obtained for all other uses, in any current or future media, including reprinting/republishing this material for advertising or promotional purposes, creating new collective works, for resale or redistribution to servers or lists, or reuse of any copyrighted component of this work in other works.
 
  Accepted for publications at DVCON-EUROPE 2025 Design and Verification Conference and Exhibition Europe}
\newcommand{\copyrightnotice}{%
\begin{tikzpicture}[remember picture,overlay]
\node[anchor=south,yshift=10pt] at (current page.south) {\fbox{\parbox{\dimexpr\textwidth-\fboxsep-\fboxrule\relax}{\copyrighttext}}};
\end{tikzpicture}%
}
\begin{document}
\bstctlcite{IEEEexample:BSTcontrol}
\title{Integrating SystemC TLM into FMI 3.0 Co-Simulations with an Open-Source Approach}
\author{Omitted for blind review}

\author{\IEEEauthorblockN{Andrei Mihai Albu$^1$, Giovanni Pollo$^1$,  Alessio Burrello$^1$, Daniele Jahier Pagliari$^1$, \\ Cristian Tesconi$^2$, Alessandra Neri$^2$, Dario Soldi$^2$, Fabio Autieri$^2$, Sara Vinco$^1$}
\IEEEauthorblockA{ 
$^1$ Politecnico di Torino, Italy - $^2$ Dumarey Group, Italy}
\IEEEauthorblockA{Emails: name.surname@polito.it}
}

\maketitle
\copyrightnotice
\begin{abstract}
 The growing complexity of cyber-physical systems, particularly in automotive applications, has increased the demand for efficient modeling and cross-domain co-simulation techniques. While SystemC Transaction-Level Modeling (TLM) enables effective hardware/software co-design, its limited interoperability with models from other engineering domains poses integration challenges. This paper presents a fully open-source methodology for integrating SystemC TLM models into Functional Mock-up Interface (FMI)-based co-simulation workflows. By encapsulating SystemC TLM components as FMI 3.0 Co-Simulation Functional Mock-up Units (FMUs), the proposed approach facilitates seamless, standardized integration across heterogeneous simulation environments. We introduce a lightweight open-source toolchain, address key technical challenges such as time synchronization and data exchange, and demonstrate the feasibility and effectiveness of the integration through representative case studies.
\end{abstract}

\section{Introduction}
\label{sec:introduction}

The increasing complexity of embedded and cyber-physical systems in modern vehicles has driven the need for advanced modeling and co-simulation techniques that enable early design exploration, performance evaluation and cross-domain integration \cite{Intro_1}. In this context, Transaction-Level Modeling (TLM) using SystemC has emerged as a powerful methodology in Electronic-System-Level (ESL) design, enabling efficient simulation and abstraction for hardware/software co-design and architectural exploration \cite{Into_3, Intro_4}.

However, modern systems are highly heterogeneous, and require to take into account multiple domains, including %
control, thermal, or mechanical \cite{Intro_5}. This limitation becomes especially critical in system-level design workflows, where co-simulation involving heterogeneous tools and models is increasingly common. As system integration becomes more multidisciplinary, the need for interoperable, modular simulation standards has become evident \cite{Modelica_book}.

To meet the growing demand for seamless integration in systems engineering, the Functional Mock-up Interface (FMI) standard has gained widespread adoption \cite{FMI_examples_1, FMI_examples_2, FMI_examples_3}. FMI enhances cross-tool interoperability, %
as it provides a standardized, tool-independent interface to %
encapsulate simulation models as %
Functional Mock-up Units (FMUs), which can be shared, reused, and integrated across diverse platforms. %

Although FMI is widely used in domains such as control systems and system dynamics (e.g., in Simulink, Dymola, or Modelica-based tools), its adoption in the SystemC TLM ecosystem remains limited.
{This work investigates and demonstrates the feasibility of integrating SystemC TLM models into FMI-based co-simulation workflows, and proposes a methodology for encapsulating SystemC TLM components as FMI 3.0 Co-Simulation FMUs. %
This integration is inherently non-intrusive, requiring no modifications to existing TLM models, and achieves full FMI standard compliance with the application of an open source automated tool. Experimental results from two industrial case studies, chosen for their diverse complexity and computation-communication tradeoffs, prove the effectiveness of our FMU integration approach. A final experiment, integrating a generated FMU with a Simulink-based FMU, conclusively proves seamless FMI compatibility.}

The rest of the paper is organized as follows: Section \ref{sec:background} presents background information on SystemC TLM, FMI standard and state of the art; Section \ref{sec:methodology} describes the proposed integration method  and its automation framework; Section \ref{sec:results} presents test studies and experimental results. Finally, Section \ref{sec:conclusions} concludes the paper and outlines future works. %

\section{Background and Related Works}
\label{sec:background}

\subsection{SystemC TLM}
SystemC TLM builds on top of the SystemC simulation engine \cite{SystemC_IEEE_1666}, by replacing low-level, pin-accurate signal protocols with higher-level transaction-based communication. This abstraction separates communication from computation, significantly reducing the complexity and improving the efficiency of simulations \cite{accellera,BACKGROUND_System_design_with_SystemC}. %

In \tlm, a \emph{transaction} represents an operation between components,  e.g., a read or write operation, rather than modeling the exact signals and timings on pins or wires.
Each transaction is implemented as a \emph{payload}, an object wrapping attributes (e.g., address, data) and protocol phases.

Transactions are exchanged between two entities: the \emph{initiator}, which triggers the operation and owns a \texttt{tlm\_initiator\_socket}, and the \emph{target}, which receives and processes the request via a \texttt{tlm\_target\_socket}. These sockets define the communication interface and support both blocking and non-blocking transport. In blocking transport, the initiator directly calls the target’s \texttt{b\_transport()} method, which processes the request immediately. In non-blocking transport, the initiator issues a request using the \texttt{nb\_transport\_fw()} method; the target acknowledges receipt, and sends the result back asynchronously via the initiator’s \texttt{nb\_transport\_bw()} method.

\subsection{Functional Mock-Up Interface}
The Functional Mock-Up Interface (FMI) %
provides a standardized application programming interface specification for the interoperability and co-simulation of dynamic models across heterogeneous simulation environments \cite{FMI_reference}. FMI facilitates cross-platform model exchange through the encapsulation of computational models %
FMUs, distributed as compressed archives containing: (1) an XML-based model description file (\texttt{modelDescription.xml}), defining the model's interface specification through structured variable declarations (e.g., direction, type, name); (2) platform-specific dynamic link libraries implementing the FMI C-API specification; and (3) optional auxiliary resource files encompassing documentation and model-specific parametric data.

FMI defines three main interactions between FMUs, i.e., Model Exchange (based on numerical integration, handled by a centralized solver), Co-Simulation (where FMUs include their solvers and execute independently), and Scheduled Execution (designed for real-time scenarios). This work focuses on Co-Simulation, that is the most frequent scenario in the context of cyber-physical systems. 

FMI 3.0 Co-Simulation specification defines a comprehensive C-API comprising mandatory and optional functions: %
\begin{itemize}
\item functions to instantiate a FMU, e.g., \texttt{fmi3InstantiateCoSimulation}; 
\item function to run simulation, e.g., \texttt{fmi3DoStep}, that allows temporal advancement to run simulation of each FMU; 
\item getter and setter functions to update FMI variables before and after each \texttt{fmi3DoStep}, to allow data exchange between FMUs: \texttt{fmi3GetXXX} and \texttt{fmi3SetXXX}\footnote{\texttt{XXX} stands for the data type, e.g., \texttt{fmi3Int8}, \texttt{fmi3Boolean}.}. 
\end{itemize}

\subsection{Related Work}
The Functional Mock-up Interface (FMI) standard has gained substantial traction in both academic research and industrial applications, finding utility across numerous domains \cite{Related_Intro}. %
Within the automotive co-simulation sector, several prominent commercial platforms have been developed to leverage FMI capabilities, such as Synopsys Silver \cite{synospss_tool}, Altair Twin \cite{Altair_tool}, Simcenter Amesim \cite{siemens_tool}, and BemNG.tech \cite{beamngBeamNGtech}. 
While these commercial solutions have achieved significant industrial penetration, they remain proprietary and generally lack native integration capabilities for SystemC models, especially those implementing SystemC TLM. 

To address this gap, multiple research efforts have investigated approaches for incorporating SystemC/SystemC TLM components into FMI-based co-simulation frameworks.
An initial contribution by \cite{Centomo_paper} presented a methodology for wrapping SystemC models as Functional Mock-up Units (FMUs) using the FMI 2.0 specification, though this approach did not address TLM-specific requirements. 
Building upon this foundation, later work \cite{Related_3} addressed the integration of TLM-level simulation, focusing on higher abstraction levels compared to RTL. Their proposed methodology begins with a VHDL or Verilog hardware description, which is then transformed into a TLM model of the IP component and subsequently wrapped into an FMU. While the process is automated, it introduces several limitations: the methodology does not directly support SystemC TLM, and an intermediate translation from RTL to TLM is required. OMSimulator \cite{OM_Simulator} represents another FMI-based co-simulation tool, tightly integrated with OpenModelica, an open-source modeling and simulation environment widely adopted in both academia and industry \cite{open_modelica}. Through this integration, OMSimulator offers a graphical interface and supports both FMI modalities: model exchange and co-simulation. Although it is build around transaction level co-simulations, it lacks native support for SystemC TLM models.

A few open source efforts are available. \cite{Related_2} introduced VPSim, that enables the automated generation of FMI-compliant FMUs through the instantiation of dedicated proxy modules for enabling communication with the system, adding a layer of complexity. Virtual Components Modeling Library (VCML) \cite{machinewareVCMLVirtual} instead offers a collection of FMU-ready SystemC TLM components for virtual prototyping with FMI support. However, {introducing new components requires modifications to the original design, and FMI support is not guaranteed.} %

To close the gap, this work aims at providing a methodology and an open source tool that automatically wraps SystemC TLM descriptions as FMUs with no modifications of the source code. The open source tool is available at \texttt{https://github.com/eml-eda/systemc-fmi}

\section{Methodology}
\label{sec:methodology}

\begin{figure}[ht]
    \centering\vspace{-0.3cm}
    \includegraphics[width=1\linewidth]{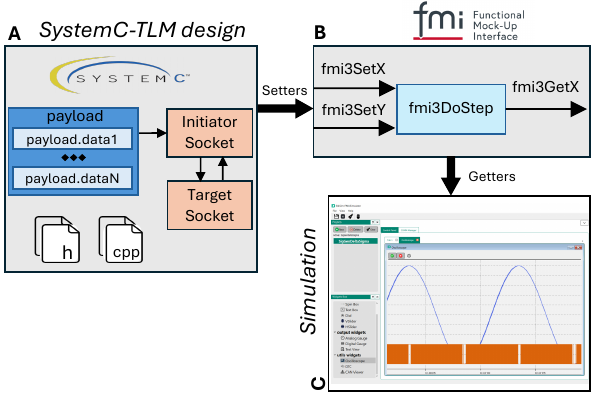}
    \caption{High-level overview of the proposed workflow}\vspace{-0.3cm}
    \label{fig:high_level_overview}
\end{figure}

This section presents a comprehensive methodology for encapsulating SystemC TLM designs as FMUs. The approach follows a structured three-phase workflow (outlined in \cref{fig:high_level_overview}): 
\begin{enumerate}[label=\emph{\Alph*.}]
    \item Design Selection and Analysis, to collect data structures and identify the communication characteristics; 
    \item FMI Wrapper Generation, to map the TLM payload to FMI variables and implement the required FMI functions; 
    \item Simulation and Validation with an FMI-compliant simulation environment (e.g. FMPy \cite{FMPy_library}). 
\end{enumerate}

\subsection{{Design Analysis and Simulation Setup}}
The methodology adopts a non-intrusive approach that preserves the integrity of existing SystemC TLM designs and imposes minimal constraints, ensuring broad applicability across different design styles.

\subsubsection{Design Prerequisites}

The methodology requires access to the SystemC TLM source code and assumes that the \tlm\ code is a target module, implementing \texttt{b\_transport} and \texttt{nb\_transport\_fw} functions. This assumption is reasonable, as TLM modules are typically used to simulate peripherals (e.g., bus interfaces), to be controlled via co-simulation by external modules (e.g., Instruction Set Simulators). Extending the methodology to support an initiator \tlm\ module would be however straightforward, and will be part of future extensions. 

\subsubsection{Design Analysis}

The code is parsed to extrapolate the necessary information. The most important element is the data format loaded in the payload as data field. This data is typically organized in a \texttt{struct}, that allows to wrap more than one variable. This \texttt{struct} is moved to an header file, \texttt{payload.h}, so that it can be accessed and imported from multiple files.

\begin{lstlisting}[
    language=C++, 
    caption={Structure loaded to payload data and saved in \texttt{payload.h}.}, 
    captionpos=b,
    label={lst:iostruct},
    basicstyle=\scriptsize\ttfamily,
    keywordstyle=\color{blue},   
    commentstyle=\color{gray},
    showstringspaces=false,
    frame=single,
    framerule=0.5pt,
    numbers=left,
    xleftmargin=3em,
    xrightmargin=1em,
    frame=single,
    framexleftmargin=2em,
    rulecolor=\color{gray!50},
    breaklines=true,
    breakatwhitespace=false,
    postbreak=\mbox{\textcolor{black}{-}\space}
]
struct payload {
    sc_dt::sc_int<32> data_in; 
    sc_dt::sc_int<32> data_out; 
};
\end{lstlisting}

Additionally, the \tlm\ functions implemented in the module are scanned to determine which of the \texttt{struct} fields are accessed through read operations (and can thus be considered as inputs received from an initiator) and which are rather set and updated in the function (and are thus outputs generated from the target).  In the example reported in \cref{lst:iostruct}, the \texttt{data\_in} field is taken in input from the target as data to elaborate, and the resulting output will be uploaded to the \texttt{data\_out} field of the \texttt{struct}. This information will be crucial to build the \texttt{ModelDescription.xml} file \cite{FMI_reference}, that 
contains all necessary metadata for the FMU.

\subsubsection{Top-Level and Initiator Module Implementation}
\tlm\ target modules can not be simulated in isolation, but rather require that their sockets are bound to TLM initiator sockets. For this reason, the methodology implements an \emph{initiator module}, in charge of handling direct communication with the target module, and a top level, that declares, instantiates and connects the initiator and the target. %

\begin{lstlisting}[
    language=C++, 
    caption={Extract of initiator module code.}, 
    captionpos=b,
    label={lst:init_block},
    basicstyle=\scriptsize\ttfamily,
    keywordstyle=\color{blue},
    commentstyle=\color{gray},
    showstringspaces=false,
    frame=single,
    framerule=0.5pt,
    numbers=left,xleftmargin=1.7em,frame=single,framexleftmargin=2em,
    rulecolor=\color{gray!50},
    breaklines=true,
    breakatwhitespace=false,
    postbreak=\mbox{\textcolor{black}{-}\space}
]
void Initiator::send_data()
{
    start_sending.notify(sc_core::SC_ZERO_TIME);
}
void Initiator::sending_thread()
{
    while(true){
        wait(start_sending);
        iostruct data_packet; 
        root_packet.data_in = data_to_send;
        tlm::tlm_generic_payload payload ;
        payload.set_command(tlm::TLM_WRITE_COMMAND);
        payload.set_data_ptr((unsigned char*) &data_packet);
        ...        
        initiator_socket->b_transport(payload, delay);
        ...
    }
}
\end{lstlisting}
\subsubsection{Initiator module} 
The initiator module declares a \texttt{tlm\_initiator\_socket} socket, used to carry as data a \texttt{struct}, as defined in the \texttt{payload.h} file. 

The initiator is in charge of starting communication with the target, via invocation of the \tlm\ primitives. This is handled with a process \texttt{sending\_thread} (lines 5-18), that repeatedly sets the payload fields and invokes the blocking or non blocking primitive of the target (an example for blocking communication is reported in \cref{lst:init_block}). The process waits on an event \texttt{start\_sending}, that is fired by a function \texttt{send\_data}, invoked from outside (as will be explained later, lines 1-4). 

\subsubsection{Top-level module} 
The top-level module is necessary to allow successful \tlm\ simulation, and \cref{lst:top} outlines its implementation. 
The constructor instantiates initiator and target and performs socket binding (lines 1-6). The top level then includes two main methods. The \texttt{set\_and\_send()} one prepares data to be transferred to the target, copies it to the initiator, and invokes the \texttt{send\_data} function of the initiator to start communication (lines 7-10). The \texttt{retrieve\_result()} is vice versa responsible for collecting the output once the transaction has completed (lines 11-13).
This solution allows controlling communication with the target and collecting data correctly through methods of the top level, without intruding in the target implementation.

\begin{lstlisting}[
    language=C++, 
    caption={Extract of the top-level module code.}, 
    captionpos=b,
    label={lst:top},
    basicstyle=\scriptsize\ttfamily,
    keywordstyle=\color{blue},
    commentstyle=\color{gray},
    showstringspaces=false,
    frame=single,
    framerule=0.5pt,
    numbers=left,
    xleftmargin=2em,
    xrightmargin=-1em,
    frame=single,
    framexleftmargin=2em,
    rulecolor=\color{gray!50},
    breaklines=true,
    breakatwhitespace=false,
    postbreak=\mbox{\textcolor{black}{-}\space}
]
Top::Top(sc_core::sc_module_name name) 
    : sc_core::sc_module(name) {
    init = new Initiator("init");
    root_ = new root("root_");
    init->initiator_socket.bind(root_->target_socket);
}
void Top::set_and_send(sc_dt::sc_int<32> data_in_to_send) {
    init->data_to_send = data_in_to_send; 
    init->send_data(); 
}
void Top::retrieve_result(sc_dt::sc_int<32> &result_out) {
    result_out = init->data_received;
}
\end{lstlisting}

\subsection{FMI Wrapper Generation}
The construction of the FMI wrapper around such system requires generating the FMU interface definition, and implementing the necessary data transfer and primitive mapping to transform FMI APIs into an evolution of the \tlm\ system.

\subsubsection{Model Description Generation}
The first step is the generation of the  \texttt{modelDescription.xml} file, which defines the FMU’s interface. Information collected about the TLM data format is used to populate the file, reporting the name of each \texttt{struct} field, annotated with the corresponding direction and type, as depicted in \cref{lst:xml} for the data structure defined in \cref{lst:iostruct}. To ensure semantic accuracy, a dedicated mapping (see Table~\ref{tab:systemc-fmi-ports}) aligns each SystemC data type with its corresponding FMI equivalent. 
\begin{lstlisting}[
    language=XML, 
    caption={Extract of the \texttt{modelDescription.xml} file \\ generated from the \texttt{struct} defined in \cref{lst:iostruct}}, 
    captionpos=b,
    label={lst:xml},
    basicstyle=\scriptsize\ttfamily,
    keywordstyle=\color{blue},
    commentstyle=\color{gray},
    showstringspaces=false,
    frame=single,
    framerule=0.5pt,
    numbers=left,xleftmargin=3em,frame=single,framexleftmargin=2em,
    rulecolor=\color{gray!50},
    breaklines=true,
    breakatwhitespace=false,
    postbreak=\mbox{\textcolor{black}{-}\space}
]
<fmiModelDescription fmiVersion="3.0" modelName="tlm"/>
  ...
  <ModelVariables>
    <Int32 name="fmi_data_in" valueReference="1" 
           causality="input" start="0"/>
    <Int32 name="fmi_result" valueReference="2" 
           causality="output"/>
  </ModelVariables>
  ...
</fmiModelDescription>
\end{lstlisting}

From listing \ref{lst:xml}, it is possible to appreciate that the \texttt{data\_in} field is mapped onto a variable \texttt{fmi\_data\_in} (with the \texttt{fmi\_} prefix) of type \texttt{Int32} (corresponding to the initial type \texttt{sc\_int<32>}). Its direction is considered as input, as it is read as input by the target. Vice versa, the \texttt{result} field is tagged as output, as its value is calculated from the target \tlm\ module, and sent back to the initiator. 

\begin{table}[htbp]
    \centering
    \caption{Data Type Mapping Between SystemC and FMI}
    \label{tab:systemc-fmi-ports}
    \resizebox{\columnwidth}{!}{
    \begin{tabular}{@{}ll@{}}
        \toprule
        \textbf{SystemC Data Type} & \textbf{FMI Data Type} \\
        \midrule
        \texttt{sc\_logic} & \texttt{fmi3Bool} \\
        \addlinespace[0.5ex]
        \texttt{sc\_bv<N>} & \texttt{fmi3Binary} \\
        \addlinespace[0.5ex]
        \multicolumn{2}{@{}l@{}}{\textit{Signed/Unsigned Integer Types:}} \\
        \quad\texttt{sc\_int<1..8>} / \texttt{sc\_uint<1..8>} & \texttt{fmi3Int8} / \texttt{fmi3UInt8}\\
        \quad\texttt{sc\_int<9..16>} / \texttt{sc\_uint<9..16>}& \texttt{fmi3Int16} / \texttt{fmi3UInt16} \\
        \quad\texttt{sc\_int<17..32>} / \texttt{sc\_uint<17..32>} & \texttt{fmi3Int32} / \texttt{fmi3UInt32} \\
        \quad\texttt{sc\_int<33..64>} / \texttt{sc\_uint<33..64>} & \texttt{fmi3Int64} / \texttt{fmi3UInt64}\\
        \addlinespace[0.5ex]
        \multicolumn{2}{@{}l@{}}{\textit{Floating Point Types:}} \\
        \quad\texttt{sc\_float} & \texttt{fmi3Float32} \\
        \quad\texttt{sc\_double} & \texttt{fmi3Float64} \\
        \bottomrule
    \end{tabular}}\vspace{-0.3cm}
\end{table}

\subsubsection{FMI Wrapper Implementation}
The next phase encapsulates the SystemC TLM model using the FMI 3.0 API. The wrapper architecture acts as a bridge between the two domains, coordinating simulation execution and managing data exchange between FMI variables and SystemC sockets. It also ensures correct propagation of inputs and collection of outputs during simulation, while enabling integration into a wider co-simulation context.
Central to this mechanism is a structured wrapper (\cref{lst:wrapper_struct}) that stores: 
\begin{itemize}
    \item a pointer \textit{*top} to the top-level \tlm\ module (\cref{lst:top}), that allows to access its data structures and to invocate its \texttt{set\_and\_send()} function; 
    \item a variable of type \texttt{sc\_time} to keep track of simulation time (for the timed versions of TLM); 
    \item the FMI interface variables defined in the XML model description (\cref{lst:xml}, lines 4-5).
\end{itemize}

\begin{lstlisting}[
    language=C++, 
    caption={SystemC-FMI wrapper struct.},
    captionpos=b,
    label={lst:wrapper_struct},
    basicstyle=\scriptsize\ttfamily,
    keywordstyle=\color{blue},
    commentstyle=\color{green!60!black},
    stringstyle=\color{red},
    showstringspaces=false,
    frame=single,
    framerule=0.5pt,
    numbers=left,xleftmargin=3em,frame=single,framexleftmargin=2em,
    rulecolor=\color{gray!50},
    breaklines=true,
    breakatwhitespace=false,
    postbreak=\mbox{\textcolor{green}{-}\space}
]
struct WRAPPER_STRUCT {
    Top *top;               // Pointer top-level module
    sc_time current_time;   // Current simulation time
    fmi3Int32 fmi_data_in;  // Input data from FMI domain
    fmi3Int32 fmi_result;   // Output data to FMI domain
};
\end{lstlisting}

\subsubsection{FMI API Implementation}
Conformance with the FMI standard requires implementing the API functions that mediate interactions between the FMI environment and  SystemC. 

The instantiation and initialization functions (i.e., \texttt{fmi3InstantiateCoSimulation} and (\texttt{fmi3EnterInitializationMode}) are used to declare and instantiate the top level entity, and to issue a \texttt{sc\_start(SC\_ZERO\_TIME)} primitive, that allows construction of the \tlm\ objects, performs the binding between sockets and initializes the event queue necessary to run the simulation. Vice versa, the \texttt{fmi3FreeInstance} function frees memory at the end. %

The setter and getter functions (\texttt{fmi3SetXXX} and \texttt{fmi3GetXXX}, \texttt{XXX} being a FMI data type) copy or retrieve the values of the FMI variables, contained in the wrapper, to local variables. The mapping onto \tlm\ payload data will be explained later on.  

The core of the cosimulation is the \texttt{fmi3doStep} function, that must advance the simulation of the \tlm\ subsystem by a certain amount of time (defined by the \texttt{CommunicationStepSize} parameter). %
This requires thus precise synchronization between the SystemC kernel and the FMI runtime, without altering core simulation behavior. An example of implementation is shown in Listing~\ref{lst:fmi3DoStep}. The function accesses the wrapper (defined in \cref{lst:wrapper_struct}) in line 7 and calculates the requested step size in seconds (line 8). The function then calls the \texttt{set\_and\_send} method of the top-level module to transfer the FMI variables into the TLM payload and to trigger the initiator. It subsequently starts the \tlm\ simulation for the specified duration using the \texttt{sc\_start} primitive (lines 11-12). The result calculated by the TLM target is retrieved with the \texttt{retrieve\_result} function of the top level, and updated to a local variable (lines 13-15). Finally, current FMU time stored in the wrapper is increased to take into account the step size just executed, considering the possibility of anticipated returns due to interrupt-like behaviors or of any error condition detected by the simulation (lines 17-21). This mechanism %
allows to temporally align the \tlm\ simulation with any other FMU, and to correctly execute its functionality. 

\begin{lstlisting}[
    language=C++, 
    caption={\texttt{fmi3DoStep} function implementation},
    captionpos=b,
    label={lst:fmi3DoStep},
    basicstyle=\scriptsize\ttfamily,
    keywordstyle=\color{blue},
    commentstyle=\color{green!60!black},
    stringstyle=\color{red},
    showstringspaces=false,
    frame=single,
    framerule=0.5pt,
    numbers=left,xleftmargin=3em,frame=single,framexleftmargin=2em,
    rulecolor=\color{gray!50},
    breaklines=true,
    breakatwhitespace=false,
    postbreak=\mbox{\textcolor{red}{$\hookrightarrow$}\space}
]
fmi3DoStep(fmi3Instance instance, 
           fmi3Float64 currentCommunicationPoint,
           fmi3Float64 communicationStepSize, ... 
           fmi3Boolean* earlyReturn,
           fmi3Float64* lastSuccessfulTime) {

    WRAPPER_STRUCT* fmu = static_cast<WRAPPER_STRUCT*>(instance);
    sc_time step_size(communicationStepSize, SC_SEC);

    fmu->top->set_and_send(static_cast<int32_t>(fmu->fmi_data_in));
    sc_start(step_size);

    int32_t result;
    fmu->top->retrieve_result(result);
    fmu->fmi_result = result;

    sc_time next_time; 
    if (!(*earlyReturn))
      next_time = fmu->current_time + step_size;
    else next_time = fmu->current_time + (*lastSuccessfulTime); 
    fmu->current_time = next_time;

    return fmi3OK;
}
\end{lstlisting}
\vspace{-0.2cm}

\subsection{FMI-based Simulation}

The compilation stage generates a platform-specific executable library (as by the FMI standard): a Dynamic Link Library \texttt{.dll} for Windows, a Shared Object Library \texttt{.so} for Linux, or a Dynamic Library \texttt{.dylib} for macOS. The compilation must ensure that all necessary dependencies are correctly linked and that the final output properly exposes the FMI interface functions required for simulation.

The final phase of the proposed workflow is simulating the generated FMU using suitable simulation tools. A typical simulation flow begins by loading the generated FMU, followed by the configuration of key parameters such as simulation duration and time step. Initial conditions and input values are then defined before executing the simulation. Once the run is complete, the results can be retrieved and analyzed as needed. For this step, the choice fell on the Python-based FMPy library~\cite{FMPy_library} for its rich feature set, intuitive interface, and ease of integration into automated environments.
Although FMPy is used as the default simulation backend in this workflow, the resulting FMUs remain fully compliant with the FMI standard. As such, they can be executed in any FMI-compatible simulation environment, making the solution broadly applicable in both academic and industrial settings.

\subsection{Automation framework}
The whole flow presented in this section has been automated 
to maximize accessibility and ease of use. %
To operate, the framework takes as input a \tlm\ model along with a configuration file (YAML or JSON), that defines the paths to source files, various FMU parameters (e.g., \texttt{CommunicationStepSize}), and other necessary metadata for the generation of the \texttt{ModelDescription.xml} file. %

The automation process is launched via a Python script that handles all subsequent stages, by following the steps detailed in the former subsections. To parse the input files (\tlm\ design, plus YAML or JSON file) we used regular expressions, and the produced code is written to files. 
Both Bash scripts and CMake are generated and supported, depending on the user’s setup. After successful compilation, the resulting binaries are packaged into a platform-specific FMU (\texttt{.dll} for Windows, a \texttt{.so} for Linux or a \texttt{.dylib} for macOS). 

By encapsulating all steps into a single automated pipeline, the framework significantly lowers the barrier for converting SystemC TLM models into FMUs suitable for co-simulation.

\vspace{-0.2cm}
\section{Experimental Results}
\label{sec:results}

To assess the effectiveness of our integration methodology, we applied the proposed flow to three case studies. {The former two are provided by an industry partner, and are used to evaluate the complexity of the wrapped model, simulation performance of native SystemC TLM simulations (including the target, the initiator, and the top level) against their FMU-based equivalents (executed via FMPy) and the peak memory consumption\footnote{The \tlm\ runs were measured using a dedicated C++ profiling framework, while FMU co-simulation was monitored in Python using \texttt{psutil}~\cite{psutil}. All measurements were averaged over five runs, with simulation lengths varying between 250 and 10,000 \texttt{DoStep} invocations.}. The two designs differ in complexity, with one focusing more on communication and the other emphasizing computation. Finally, seamless FMI compatibility is demonstrated through the successful integration of a generated FMU with a Simulink-based FMU. }

\subsection{I2C (Inter Integrated Circuit)}
\label{subsec: I2C (Inter Integrated Circuit)}
The first case study includes an I2C bus, receiving requests from the FMI interface and handling communication with two slaves: an Arithmetic Logic Unit (ALU) and a Register File. %
The I2C description comprises a master controller, orchestrating protocol-level operations, e.g., address transmission and acknowledgments, and a slave interface, managing the I2C protocol state machine. %
The resulting \tlm\ includes thus 4 modules for a total of 5 processes and 1,072 lines of code. Total payload data includes 7 fields (3 booleans, 2 \texttt{uint8}, and 2 enumerative types). 

\begin{table}[ht]
    \centering
    \caption{I2C performance as execution time and memory footprint when increasing the number of \texttt{fmi3doStep} invocations}
    \label{tab:result:i2c}
    \begin{tabular}{|c|r|r|r|r|r|r|}
    \hline
         \rowcolor{Gainsboro!60}\textbf{\texttt{doStep} (\#)}& \multicolumn{2}{c}{\textbf{250}}  & \multicolumn{2}{|c}{\textbf{1,000}} & \multicolumn{2}{|c|}{\textbf{10,000}} \\    \hline\hline
         Version& TLM & FMU & TLM & FMU & TLM & FMU \\
         \hline\hline
         Time (ms)& 45.44 & 46.50 &64.93 & 413.00 & 370.94 & 3,750.00 \\
         \hline
         Memory (MB)& 5.40 & 78.40 & 5.28 & 79.27 & 6.30 & 80.85\\
         \hline
    \end{tabular}
\end{table}

Table \ref{tab:result:i2c} reports the evolution of simulation time and memory footprint when increasing the number of \texttt{fmi3doStep} invocations. 
The overhead ranges from approximately 1.02$\times$ with 250 invocations of the \texttt{fmi3doStep} function to around 10$\times$ with 10,000 invocations. This performance degradation can be attributed to the introduction of additional interfacing layers, resulting in computational and communication overhead, plus to the overhead of time synchronization, data serialization, and inter-process communication. On the contrary, the native SystemC TLM implementation benefits from direct access to efficient time management and transaction-level abstractions, which are partially lost when the module is encapsulated as an FMU. 

However, such overhead is reasonable when considering that this allows to cosimulate the wrapped TLM with external tools. As an example, \cref{fig:high_level_overview}.C shows a wrapper \tlm\ FMU executed and controlled by a Software-in-the-Loop (SIL) environment,  used to execute control software interacting with emulated hardware and peripherals, for automotive early virtual prototyping \cite{silsim}. 

Memory consumption increases in the FMU-based implementation. For instance, at 250 invocations, the FMU uses 78.4 MB, compared to just 5.4 MB in the SystemC TLM case. This pattern is similar at higher invocation counts, with FMU memory usage being around 80.85 MB at 10,000 invocations, while the native version uses 6.3 MB. The increased memory usage in the FMU version is primarily due to the additional runtime environment and infrastructure required to support the co-simulation interface, and is quite stable when increasing simulation length (with only a 3\% increase from 250 to 10,000 \texttt{fmi3doStep} invocations). 

\subsection{ECC (Error Correction Code)}
The Error Correction Code (ECC) is an crucial component for automotive systems to ensure robust error detection and correction in harsh and electromagnetically noisy environments \cite{ECC-example}. The ECC includes XOR-based parity check, dynamic adjustment based on system configuration, and support for both byte mode (8-bit) or word mode (16-bit). The \tlm\ target includes only 1 module with 3 processes, for overall 1,311 lines of code. Total payload data includes 7 fields (7 logic values and 3 logic vectors).  

\begin{table}[t]
    \centering
    \caption{ECC performance as execution time and memory footprint when increasing the number of \texttt{fmi3doStep} invocations}
    \label{tab:result:ecc}
    \begin{tabular}{|c|r|r|r|r|r|r|}
    \hline
         \rowcolor{Gainsboro!60}\textbf{\texttt{doStep} (\#)}& \multicolumn{2}{c}{\textbf{250}}  & \multicolumn{2}{|c}{\textbf{1,000}} & \multicolumn{2}{|c|}{\textbf{10,000}} \\    \hline\hline
         Version& TLM & FMU & TLM & FMU & TLM & FMU \\
         \hline\hline
         Time (ms)& 77.72 & 88.30 & 170.11 & 373.90 & 1,185 & 4,945 \\
         \hline
         Memory (MB)& 4.56 & 104.06 & 4.72 & 105.55 & 5.82 & 124.85\\
         \hline
    \end{tabular}\vspace{-0.5cm}
\end{table}

The ECC module exhibits a similar performance overhead pattern to the I2C module when encapsulated as an FMU (see \cref {tab:result:ecc}). Execution time degradation ranges from approximately 1.36$\times$ for a simulation with 250 invocations of the \texttt{fmi3doStep} function to 3.17$\times$ for 10,000 invocations. The higher overhead observed at 250 can be attributed to the fixed costs of FMU encapsulation being less effectively amortized over a shorter simulation. In contrast, this overhead becomes proportionally smaller in longer simulations.

Memory usage patterns follow a trend similar to the I2C implementation, with an almost constant overhead of 22$\times$. The higher overhead w.r.t. the I2C design (in avg. 14$\times$) is due to the computation-oriented nature of the ECC design, that thus requires more data transfers and a higher memory occupancy. 

\subsection{Model-Based Development: SystemC-TLM and Multi-Domain FMU Co-simulation}
\begin{figure}[!t]
    \centering\vspace{-0.3cm}
    \includegraphics[width=1\linewidth]{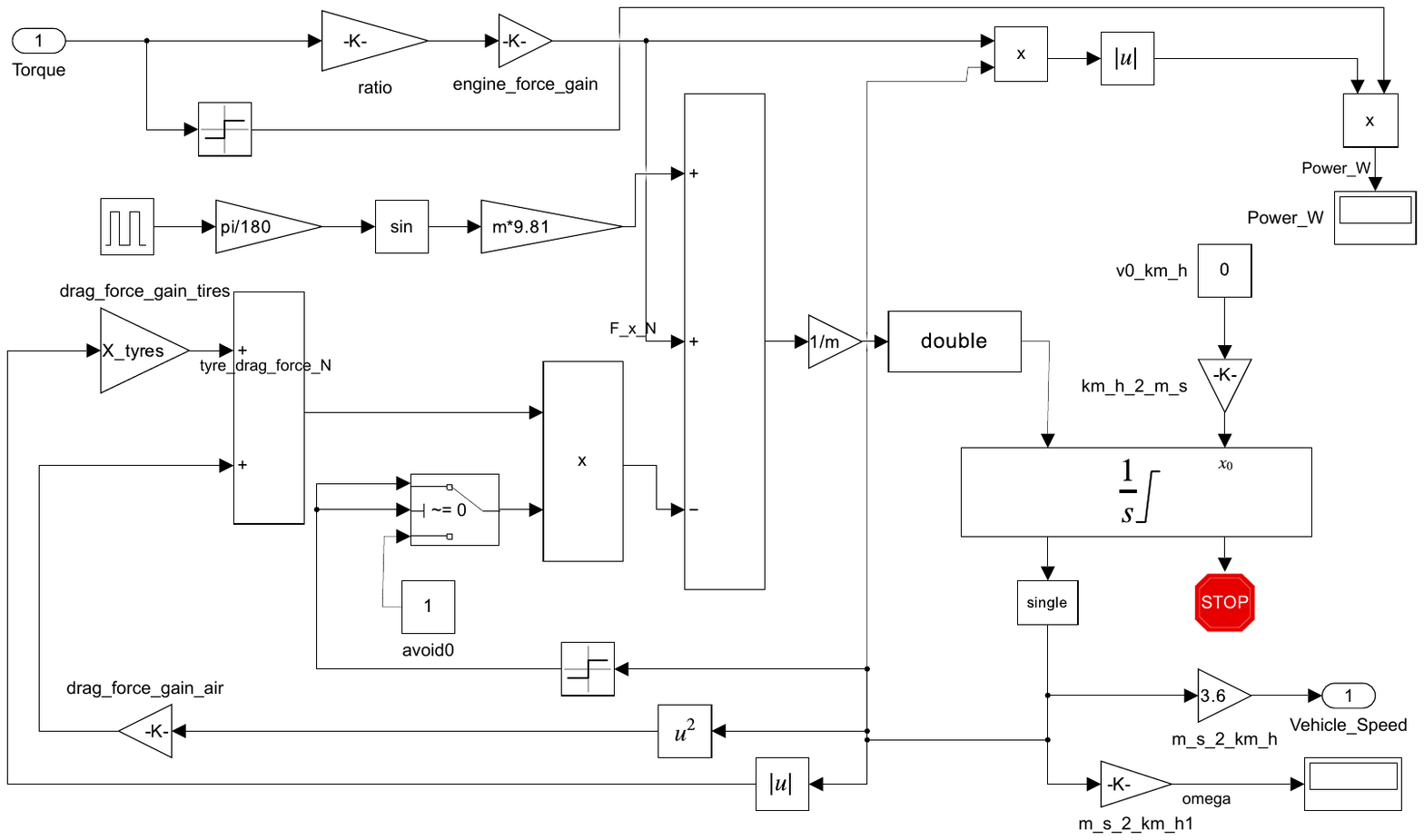}\vspace{-1cm}
    \caption{Simulink model of a single-pedal electric vehicle}\vspace{-0.5cm}
    \label{fig:simulink}
\end{figure}

The last case study focuses on the interoperability of the generated FMUs with FMUs developed in other modeling environments. 
As case study, we selected a single-pedal electric vehicle implemented in Simulink (represented in \cref{fig:simulink}). 
The modeled vehicle features a 1,600 kg mass, 90 horsepower drivetrain, and a unified pedal interface that provides both acceleration and regenerative braking functionality. The exported Simulink FMU encapsulates the complete vehicle dynamics through 19 variables: time, input torque request (Nm), output vehicle speed (km/h), and 16 configurable parameters defining the vehicle characteristics. The Simulink design was exported as an FMU using the CATIA FMI-Kit \cite{Sommer2024CATIA}. To complement this plant model, a \tlm\ Electronic Control Unit (ECU) was developed to generate representative driving scenarios. The ECU outputs the commanded torque values to be fed to the vehicle plant. 

The \tlm\ design was packaged as an FMU with the approach proposed in this work. 
The co-simulation was executed using the FMPy library to orchestrate the interaction between both FMUs. 

\begin{figure}[!th]
    \centering\vspace{-0.3cm}
    \includegraphics[width=1\linewidth]{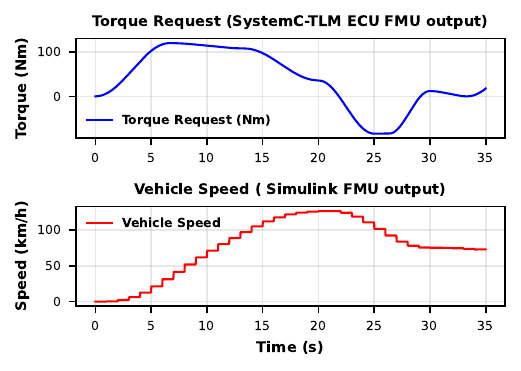}\vspace{-0.3cm}
    \caption{Cosimulation of the \tlm\ ECU with the Simulink model in \cref{fig:simulink}: torque request (top) and vehicle speed (bottom).}\vspace{-0.5cm}
    \label{fig:ECU_SLX}
\end{figure}

Simulation results (reported in \cref{fig:ECU_SLX}) demonstrate successful coupling between the \tlm\ ECU (top) and the Simulink vehicle model (bottom) over a 35-second test scenario. The torque request profile generated by the ECU exhibits a characteristic driving cycle, beginning with gradual acceleration (0-120 Nm), maintaining steady-state operation, followed by regenerative braking with negative torque values (down to -80 Nm), and concluding with a return to cruise conditions. Vehicle speed response shows appropriate dynamic behavior, accelerating from rest to approximately 120 km/h during positive torque phases and decelerating during regenerative braking periods. The response exhibits realistic vehicle inertia characteristics, with smooth speed transitions that follow the torque command profile with expected delays inherent to vehicle dynamics. 

The piecewise constant curve of the Simulink output (bottom) is due to the different time steps used by the two cosimulations (0.1s for \tlm, 1s for Simulink). This confirms that the time synchronization mechanism is robust, as it allows correct cosimulation with FMUs running with a different \texttt{CommunicationStepSize} setting. 

This successful co-simulation confirms that the FMUs generated by our framework are fully compliant with the FMI standard and are designed to be modular and interoperable. As such, they can be seamlessly integrated with FMUs developed using other tools and modeling environments, enabling flexible and multi-domain simulation workflows.

\section{Conclusions}
\label{sec:conclusions}
{In this paper, we presented an open-source automated framework for integrating SystemC TLM models with the FMI standard, facilitating co-simulation and cross-domain interoperability. 
The framework and the accompanying open source tool automatically generate FMUs from unmodified \tlm\ models, ensuring tool-independent and non-intrusive integration. Experimental results demonstrate its ability to handle a wide range of designs efficiently, with consistent memory usage and moderate performance overhead. Additionally, the experimental results prove the straightforward integration with other FMUs. Future work will focus on extending support for additional SystemC TLM features and integrating with Instruction Set Simulator (ISS) based environments.}

\newpage
\bibliographystyle{IEEEtran}
\bibliography{references}

\begin{thebibliography}{10}
\providecommand{\url}[1]{#1}
\csname url@samestyle\endcsname
\providecommand{\newblock}{\relax}
\providecommand{\bibinfo}[2]{#2}
\providecommand{\BIBentrySTDinterwordspacing}{\spaceskip=0pt\relax}
\providecommand{\BIBentryALTinterwordstretchfactor}{4}
\providecommand{\BIBentryALTinterwordspacing}{\spaceskip=\fontdimen2\font plus
\BIBentryALTinterwordstretchfactor\fontdimen3\font minus \fontdimen4\font\relax}
\providecommand{\BIBforeignlanguage}[2]{{%
\expandafter\ifx\csname l@#1\endcsname\relax
\typeout{** WARNING: IEEEtran.bst: No hyphenation pattern has been}%
\typeout{** loaded for the language `#1'. Using the pattern for}%
\typeout{** the default language instead.}%
\else
\language=\csname l@#1\endcsname
\fi
#2}}
\providecommand{\BIBdecl}{\relax}
\BIBdecl

\bibitem{Intro_1}
Z.~Zhang \emph{et~al.}, ``Co-simulation framework for design of time-triggered cyber physical systems,'' \emph{Simulation Modelling Practice and Theory}, vol.~43, p. 16–33, 04 2014.

\bibitem{Into_3}
G.~Martin, ``Embedded system design: Modeling, synthesis, and verification,'' \emph{Design \& Test of Computers, IEEE}, vol.~27, pp. 82 -- 83, 05 2010.

\bibitem{Intro_4}
G.~Martin and G.~Smith, ``High-level synthesis: Past, present, and future,'' \emph{IEEE Design \& Test of Computers}, vol.~26, no.~4, pp. 18--25, 2009.

\bibitem{Intro_5}
A.~Mahmoudi \emph{et~al.}, ``A systematic mapping study on {SystemC/TLM} modeling capabilities in new research domains,'' \emph{ACM Transactions on Design Automation of Electronic Systems}, vol.~30, no.~4, Jun. 2025.

\bibitem{Modelica_book}
P.~A. Fritzson, \emph{\BIBforeignlanguage{en}{Principles of object-oriented modeling and simulation with modelica 3.3}}, 2nd~ed.\hskip 1em plus 0.5em minus 0.4em\relax Nashville, TN: John Wiley \& Sons, Nov. 2014.

\bibitem{FMI_examples_1}
E.~Widl \emph{et~al.}, ``{FMI}-based co-simulation of hybrid closed-loop control system models,'' in \emph{Proc. of ICCSE}, 2015, pp. 1--6.

\bibitem{FMI_examples_2}
A.~Haider \emph{et~al.}, ``Modeling and simulation of automotive {FMCW RADAR} sensor for environmental perception,'' \emph{IEEE Open Journal of Intelligent Transportation Systems}, vol.~6, pp. 433--455, 2025.

\bibitem{FMI_examples_3}
F.~Perabo and M.~Zadeh, ``Multiphysics modeling and co-simulation of ship electric power and propulsion systems for virtual testing and verification,'' \emph{IEEE Transactions on Transportation Electrification}, vol.~11, no.~1, pp. 5108--5121, 2025.

\bibitem{SystemC_IEEE_1666}
``{IEEE Standard for Standard SystemC Language Reference Manual},'' \emph{IEEE Std 1666-2023 (Revision of IEEE Std 1666-2011)}, pp. 1--618, 2023.

\bibitem{accellera}
``accellera.org,'' \url{https://www.accellera.org}, July 2009.

\bibitem{BACKGROUND_System_design_with_SystemC}
T.~Grötker \emph{et~al.}, \emph{System Design with SystemC}.\hskip 1em plus 0.5em minus 0.4em\relax Springer, 01 2002.

\bibitem{FMI_reference}
Modelica, ``{Modelica/FMI-standard}: Specification of the functional mock-up interface (fmi),'' \url{https://fmi-standard.org/}.

\bibitem{Related_Intro}
S.~Hansen \emph{et~al.}, ``The {FMI 3.0} standard interface for clocked and scheduled simulations,'' \emph{Electronics}, vol.~11, p. 3635, 11 2022.

\bibitem{synospss_tool}
Synopsys, ``Silver: Software in the loop for virtual ecus | synopsys,'' \url{https://www.synopsys.com/verification/virtual-prototyping/silver.html#features}, 2025.

\bibitem{Altair_tool}
Altair, ``Altair twin,'' \url{https://altair.com/twin-activate}, 2025.

\bibitem{siemens_tool}
Siemens, ``Simcenter amesim,'' \url{https://plm.sw.siemens.com/en-US/simcenter/systems-simulation/amesim/}, 2025.

\bibitem{beamngBeamNGtech}
BeamNG, ``Beamng.tech,'' \url{https://beamng.tech/}, 2025.

\bibitem{Centomo_paper}
S.~Centomo \emph{et~al.}, ``Using {SystemC} cyber models in an {FMI} co-simulation environment: Results and proposed {FMI} enhancements,'' in \emph{Proc. of Euromicro DSD}, 2016, pp. 318--325.

\bibitem{Related_3}
------, ``Transaction-level functional mockup units for cyber-physical virtual platforms,'' in \emph{Proc. of FDL}, 2018, pp. 5--8.

\bibitem{OM_Simulator}
L.~Ochel \emph{et~al.}, ``{OMSimulator - Integrated FMI and TLM-based Co-simulation with Composite Model Editing and SSP},'' in \emph{International Modelica Conference}, 02 2019, pp. 69--78.

\bibitem{open_modelica}
{Open Source Modelica Consortium}, ``Openmodelica,'' \url{https://openmodelica.org/}.

\bibitem{Related_2}
S.~E. Saidi \emph{et~al.}, ``Fast virtual prototyping of cyber-physical systems using {SystemC and FMI: ADAS} use case,'' 10 2019, pp. 43--49.

\bibitem{machinewareVCMLVirtual}
``{V}{C}{M}{L} {V}irtual {P}latform: {O}pen-{S}ource {S}ystem{C} {T}{L}{M}-2.0 {L}ibrary - machine-ware.de --- machineware.de,'' \url{https://www.machineware.de/products/vcml-virtual-platform}, 2025.

\bibitem{FMPy_library}
``{CATIA-Systems/FMPy}: Simulate functional mockup units ({FMUs}) in python,'' \url{https://github.com/CATIA-Systems/FMPy}.

\bibitem{psutil}
\BIBentryALTinterwordspacing
psutil, ``psutil.'' [Online]. Available: \url{https://github.com/giampaolo/psutil}
\BIBentrySTDinterwordspacing

\bibitem{silsim}
D.~Group, ``Silsim,'' \url{https://www.dumarey.com/solution/virtual-prototyping/}, 2025.

\bibitem{ECC-example}
A.~Tadimarri \emph{et~al.}, ``Driving towards safety: The role of {ECUs and IMUs} in advanced driver-assistance systems {(ADAS)},'' \emph{International Journal For Multidisciplinary Research}, vol.~6, 04 2024.

\bibitem{Sommer2024CATIA}
T.~Sommer \emph{et~al.}, ``Catia-{Systems}/{FMIKit}-{Simulink},'' \url{https://github.com/CATIA-Systems/FMIKit-Simulink}, 2024.

\end{thebibliography}

\end{document}